# Taramandal-GPT: Enhancing Astrodynamics Problem-Solving with Knowledge Retrieval and Structured Thinking

Akhil Sharma, Jatin Gupta, Ali Imam Abidi*
Department of Computer Science and Engineering
Sharda University
sharmaakhil944@gmail.com, jatingupta261001@gmail.com, aliabidi4685@gmail.com

**Abstract**

***Large language models (LLMs) have shown remarkable progress in natural language understanding, yet their effectiveness in specialized fields like astronomy and astrodynamics remains limited due to challenges in multi-step reasoning, symbolic manipulation, and domain-specific terminology. To address this, we present Taramandal-GPT (Constellation-GPT), a domain-adapted framework built on the Qwen3-8b backbone, enhanced with a Retrieval- Augmented Generation (RAG) pipeline and a fallback mechanism for improved contextual precision. We evaluate it on the Astrodynamics Problems Benchmark (APBench), a dataset of 299 questions covering foundational to advanced levels of space science. Using a dual evaluation method—numeric margin - based scoring and semantic similarity assessment—Taramandal- GPT achieves competitive performance against state-of-the-art open- and closed-source models, with notable strength in thinking-intensive tasks. These results highlight the value of specialized LLMs for domains demanding accuracy and interpretability, positioning Taramandal-GPT as a step toward reliable Artificial Intelligence (AI) assistants for astrophysics, spacecraft engineering, and space exploration.***

## 1. Introduction

The emergence of Large Language Models (LLMs), beginning with Vaswani et al.'s "Attention Is All You Need", has transformed natural language processing through the Transformer architecture [1]. These models generalize well across domains and can extract and synthesize knowledge effectively. In astronomy, LLMs perform adequately in basic factual recall but struggle with complex tasks such as interpreting terminology, applying physical laws, and solving computational problems, revealing limitations in mathematical reasoning and domain-specific expertise [2].

Nonetheless, interest in applying LLMs to scientific and engineering contexts is growing. For example, the German Space Operations Center (GSOC) has explored their use in spacecraft engineering for real-time troubleshooting, documentation management, and decision support [3]. In parallel, domain-focused models such as AstroLLaMA [4] and astroBERT [5] show improved factual accuracy, contextual reasoning, and literature processing, demonstrating the value of targeted fine-tuning.

Yet, a key gap remains: no specialized thinking-focused LLM has been developed for astrodynamics and astrophysics, where both conceptual understanding and mathematical rigor are essential. To address this, we propose Taramandal-GPT (Constellation-GPT), a domain-adapted framework that integrates retrieval-based contextual grounding with structured thinking mechanisms. By emphasizing factual accuracy and symbolic computation, our approach seeks to move beyond general-purpose assistance toward reliable tools for advancing space science and engineering.

## 2. Related Works

Recent advances in domain-specific AI for astronomy have produced several notable models, including astroBERT [7], a 110M-parameter BERT-based model trained on astronomical literature for semantic search and entity recognition; the AstroLLaMA family, with AstroLLaMA27B achieving a 30% perplexity reduction over LLaMA2 after fine-tuning on 300k astronomy abstracts [8], and AstroLLaMA38B extending training with OCR-processed and summarized paper sections [9]; the AstroMLab project's AstroSage-LLaMA 3.1-8B, reaching 80.9% accuracy on AstroMLab-1 and matching GPT-4o on fewer parameters [6], and the larger AstroSage-LLaMA3.1-70B, tying with Claude-4Opus at 86.2% [7]; the StarWhisper LightCurve series for stellar classification, with a Swin Transformer variant attaining 99% accuracy [8]; These developments collectively underscore the rapid maturation of AI-driven tools as indispensable assets for modern astronomical research.

## 3. Methodology

This section outlines the methodology adopted for developing Taramandal -GPT, including dataset preparation, model selection, prompt engineering, and the design of the retrieval- augmented generation pipeline with a fallback mechanism.

### 3.1 Dataset Description

For model tuning, we drew on open-source materials and books such as Fundamentals of Physics [9], Introduction to Space Physics [10], High-Energy Astrophysics [10], and related references. Then, the text was extracted and was converted in a markdown strctutred format using the PymuPDF library. The text was embedded using the Qwen3-Embedding-0.6B model [12] due to its exceptional multilingual capabilities, strong long-text understanding, efficient 0.6B parameter size, and state-of-the-art performance across diverse text embedding and retrieval tasks.

For testing the full potential of LLMs in astrodynamics, we have utilized the first **Astrodynamics Problems Benchmark (APBench)** [13] to evaluate the capabilities of LLMs in this field. The benchmark consists of 299 QA questions drawn from authoritative aerospace engineering sources, spanning difficulty from foundational concepts to PhD-level problem-solving.

Table 1: Level-wise Distribution of APBench Questions

| **Benchmark Subset** | **Question Count** |
|---|---|
| APBench-α | gordon (17) + UBC (8) |
| APBench-β | α (25)+ Braeunig (144) |
| APBench-γ | β (169) + lynnane (130) |
| Total | 299 |

The dataset is structured into three sequential levels—alpha (α), beta (β), and gamma (γ)—with their respective proportions presented in Table 1.

### 3.2 Proposed Framework

The proposed framework, Taramandal-GPT (depicted in Figure 2), incorporates a base pre-trained model, a system prompt, and an associated pipeline, each of which is described in the following subsections.

#### 3.1.1. Base Model

This study employs Qwen3:8b [14], an open-source large language model renowned for its reasoning capabilities in complex calculations and knowledge- intensive tasks, and competes with other models. The model was deployed with Q_4-bit quantization to enable lightweight execution and efficient accessibility.

#### 3.1.2. System Prompt

The system prompt integrates chain-of-thought reasoning, agent identity utilization, and regulatory mechanisms. These features work together to fine-tune the model for domain-specific responses, thereby enhancing accuracy. An overview of the system prompt is presented in Figure 1.

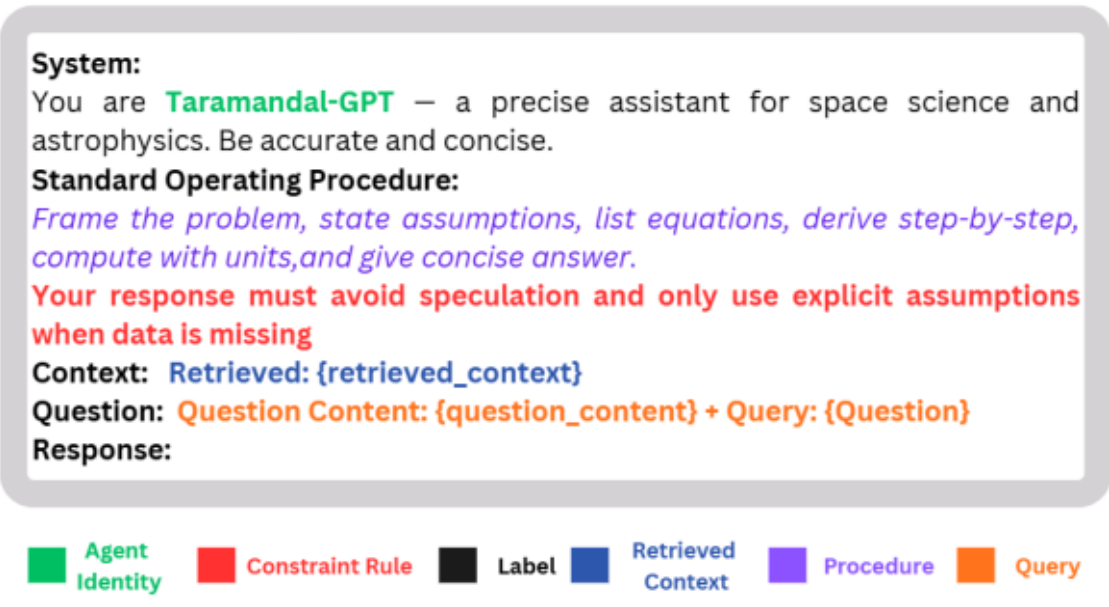

*Figure 1: Structure of System Prompt*

### 3.1.3. Framework Pipeline

The proposed framework, Taramandal-GPT, employs a Retrieval-Augmented Generation (RAG)- based architecture enhanced with a fallback mechanism. In the primary workflow, the system retrieves relevant contextual information from an external knowledge source and integrates it with the model's generative capabilities to produce precise, context-aware responses. However, when the retrieved context is insufficient or incomplete for generating a reliable output, the fallback mechanism is activated. In this mode, the model leverages its pre-trained knowledge base and engages in reasoning processes to formulate an accurate and coherent response, even in the absence of adequate external data.

The overall architectural design, encompassing both the RAG pipeline and the fallback pathway, is illustrated in Figure 2, providing a clear visual representation of the system's operational flow and decision logic.

## 4. Results and Discussion

### 4.1. Model Evaluation Metric

For APBench, model performance is assessed using a dual-format scoring schema based on the nature of the expected output: numeric or message-based.

#### 4.1.1. Numeric Answer Scoring

Numeric responses were evaluated against ground truth using a dynamic error margin, which is defined in Equation 1.

$$\textbf{Error_Margin} = \min(0.1 + 0.01 \cdot \log(|\text{Answer}|), 0.1)$$

A model's output is considered correct if it falls within this margin. For instance, if the true answer is −5.2, the acceptable range would be approximately [−5.655, −4.479].

#### 4.1.2. Message Answer Scoring

Message-based responses are evaluated using a hybrid similarity approach that combines both semantic judgment and embedding-based comparison. First, GPT-4o serves as an LLM-as-a-Judge, assigning a similarity score between 0 and 10 based on the alignment between the model-generated response and the reference answer; this score is then normalized to a [0, 1] scale. In parallel, cosine similarity is computed using sentence embeddings derived from the all-MiniLM-L6-v2 model. The final score is obtained by averaging these two components, as per Equation 2.

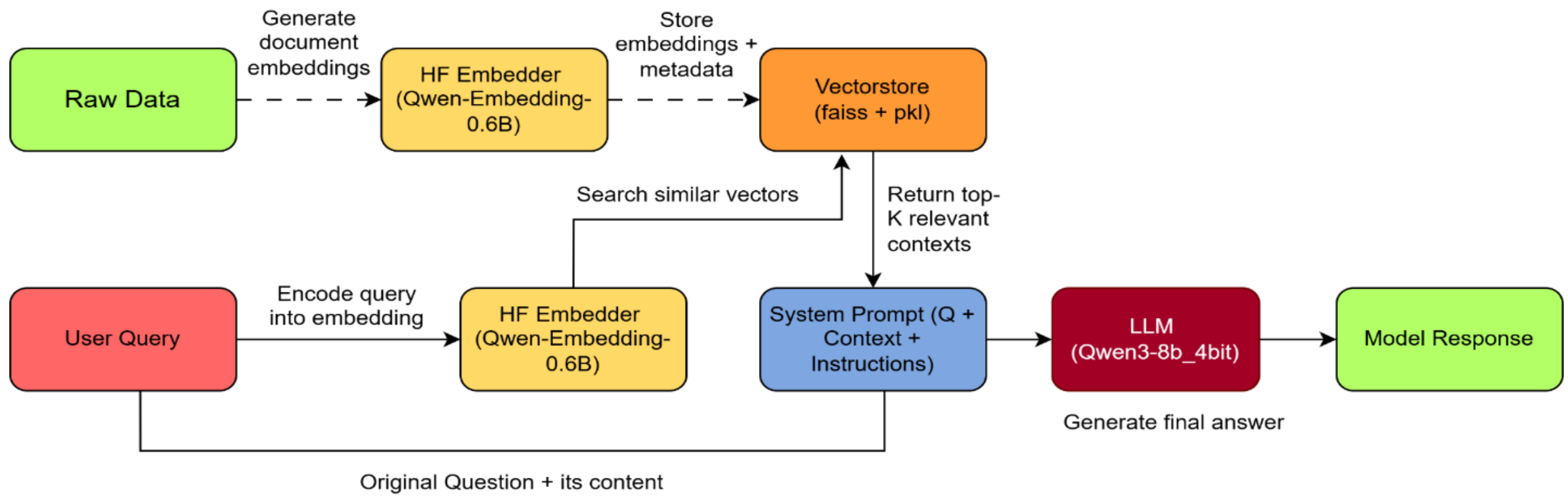

*Figure 2: Taramandal-GPT architecture with RAG pipeline & fallback mechanism*

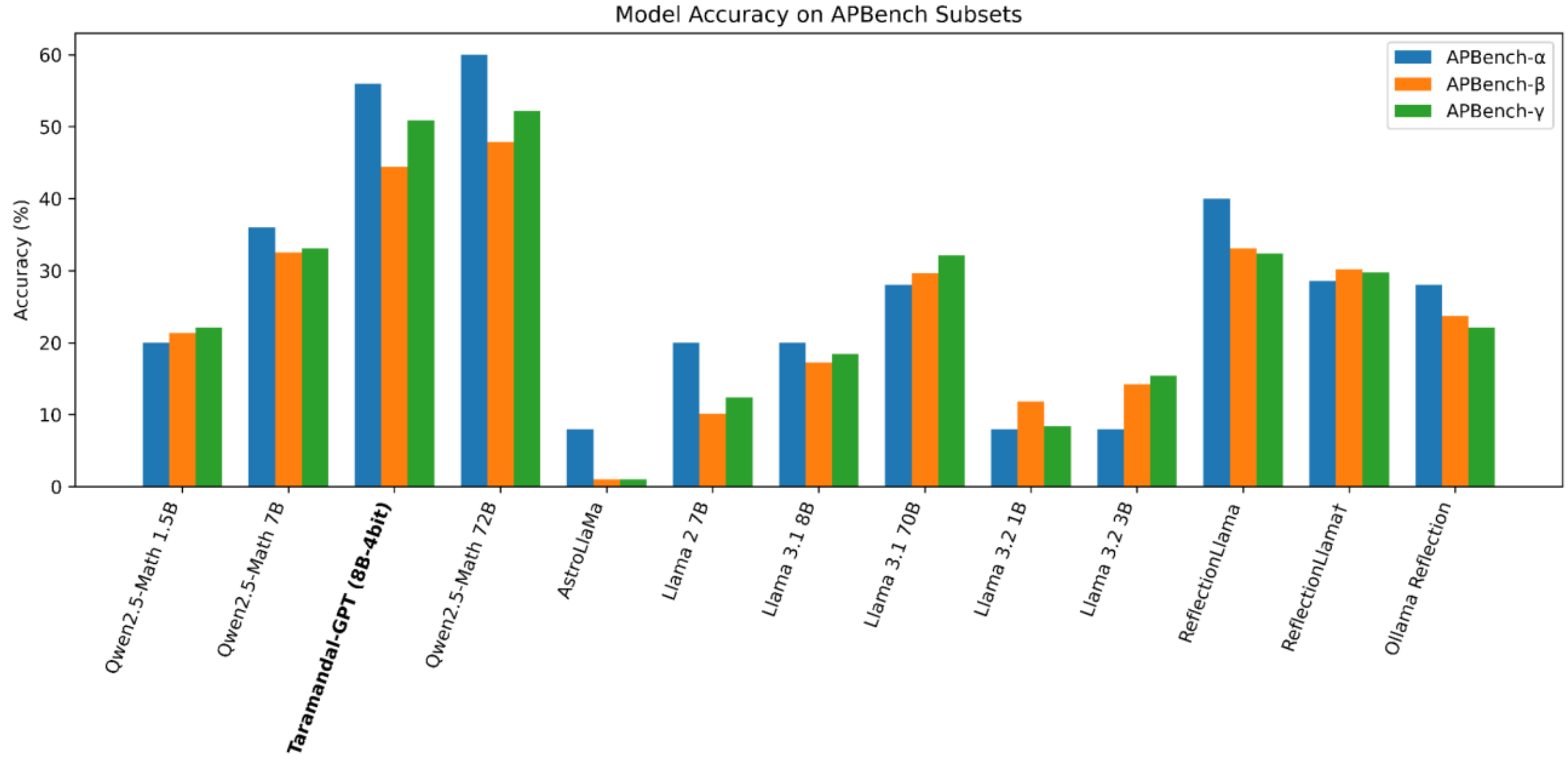

*Figure 3: Performance comparison of Taramandal-GPT with closed- and open-source models on APBench-α, β, and γ.*

**Score = 1/2 × (LLM-Judge + Embedding Similarity)**

Responses achieving a score of 0.6 or higher are considered accurate within the benchmark evaluation framework.

### 4.2 Evaluation Protocol

Zero-shot prompting is employed throughout the evaluation process, wherein models are presented with questions and context without any prior examples or demonstrations.

### 4.3 Performance of Taramandal- GPT

The performance of the proposed framework, Taramandal-GPT, is summarized in Table 2

*Table 2: Performance of Taramandal- GPT on different levels of APBench*

| *APBench* | *α* | *β* | *γ* |
|---|---|---|---|
| *Accuracy (%)* | **56** | **44.37** | **50.84** |

It scores 56% on APBench-α, 44.37% on β, and 50.84% on γ, showing competitive performance despite its smaller size when compared to much larger models like Qwen2.5-Math 72B.

### 4.4 Comparative Performance of Taramandal-GPT vs Peer Models on APBench

We compare performance of Taramandal-GPT with other closed and open-source models on APBench-α, APBench-β, APBench-γ. Figure 3 visualizes the model-wise performance trends extracted from the detailed evaluation results reported in Table 3.

The models are grouped as closed source models and open source models. Despite of the individual model's performance variation on the three Open APBench datasets, the difference between closed source models' performance and open source models' performance is clear.

Table 3: LLMs Performance on APBench

| Model | APBench-α (%) | APBench-β (%) | APBench-γ (%) |
|---|---|---|---|
| Qwen2.5-Math 1.5B | 20 | 21.3 | 22.1 |
| Qwen2.5-Math 7B | 36 | 32.5 | 33.1 |
| **Taramandal-GPT** | **56** | **44.37** | **50.84** |
| Qwen2.5-Math 72B | 60 | 47.9 | 52.2 |
| AstroLlaMa | 8 | 1 | 1 |
| Llama 2 7B | 20 | 10.1 | 12.4 |
| Llama 3.1 8B | 20 | 17.2 | 18.4 |
| Llama 3.1 70B | 28 | 29.6 | 32.1 |
| Llama 3.2 1B | 8 | 11.8 | 8.4 |
| Llama 3.2 3B | 8 | 14.2 | 15.4 |
| ReflectionLlama | 40 | 33.1 | 32.4 |
| ReflectionLlama† | 28.6 | 30.2 | 29.8 |
| Ollama Reflection | 28 | 23.7 | 22.1 |

## 5. Conclusion & Future Scope

In this work, we introduced Taramandal-GPT, a reasoning- specialized large language model framework for advancing problem-solving in astronomy and astrodynamics. By combining a Retrieval-Augmented Generation (RAG) pipeline with a fallback mechanism, it improves contextual awareness and generates more reliable responses than general-purpose and astronomy- focused models. Evaluation on APBench, a benchmark for astrodynamics problems, shows its ability to handle tasks from foundational principles to advanced research scenarios. These results highlight the potential of specialized LLMs in scientific domains where precision, mathematical rigor, and interpretability are essential, though challenges such as hallucinations and slow deliberation remain.

Looking ahead, future work includes expanding the training corpus with peer-reviewed literature, spacecraft telemetry, and mission design data to improve factual grounding; integrating neuro-symbolic and physics-informed methods for greater mathematical reliability; and developing agent-based extensions to support interactive collaboration with scientists and engineers. Broader benchmarking across domains such as planetary science, satellite communication, and exoplanetary modeling will further test the framework's robustness and applicability.

# तारामंडल-GPT: खगोलगतिकी में ज्ञान पुनर्प्राप्ति व संरचित चिंतन द्वारा समस्या-समाधान

अखिल शर्मा, जतिन गुप्ता एवं अली इमाम आबिदी
शारदा विश्वविद्यालय, ग्रेटर नोएडा, उत्तर प्रदेश
sharmaakhil944@gmail.com, jatingupta261001@gmail.com, aliabidi4685@gmail.com

**सारांश - पिछले कुछ समय में Large Language Models (LLMs) ने प्राकृतिक भाषा को समझने में जबरदस्त प्रगति दिखाई है, लेकिन खगोलशास्त्र और खगोलगतिकी जैसे खास क्षेत्रों में इनकी प्रभावशीलता सीमित रही है। इनका मुख्य कारण हैं बहुचरणीय सोच, प्रतीकात्मक गणना और क्षेत्र-विशेष शब्दावली द्वारा उत्पन्न होने वाली चुनौतियाँ। इसे हल करने के लिए, हम प्रस्तुत करते हैं तारामंडल-GPT। यह Qwen3-8b पर आधारित एक क्षेत्र-अनुकूलित मॉडल है, जिसमें 'Retrieval Augmented Generation (RAG)' प्रणाली और सटीकता के लिए एक 'Fallback' सुविधा जोड़ी गई है। हमने इसका मूल्यांकन Astrodynamics Problems Benchmark (APBench) पर किया, जिसमें 299 सवाल शामिल हैं, जो अंतरिक्ष विज्ञान के मूलभूत से लेकर उन्नत स्तर तक के विषयों से जुड़े हुए हैं। दोहरी मूल्यांकन विधि, 'numeric margin' आधारित 'scoring' और 'semantic similarity' का उपयोग करते हुए, तारामंडल-GPT ने विचार-प्रधान कार्यों में उत्कृष्टता के साथ अन्य आधुनिक खुला स्रोत मॉडल्स के अपेक्षाकृत बेहतर प्रदर्शन किया। ये परिणाम साबित करते हैं कि विशेषतः सटीकता और व्याख्या की आवश्यकता वाले क्षेत्रों में LLMs कितने उपयोगी हैं। तारामंडल-GPT खगोलभौतिकी, अंतरिक्ष यान अभियांत्रिकी और अंतरिक्ष में खोज के लिए भरोसेमंद AI असिस्टेंट की दिशा में एक महत्वपूर्ण कदम है।**

## 1. प्रस्तावना

Vaswani et al. के "Attention Is All You Need" [1] पेपर के साथ Large Language Models (LLMs) का पदार्पण हुआ, जिससे natural language processing (NLP) में बहुत क्रांति आई है, खासकर Transformer architecture की वजह से। ये models अलग-अलग विधाओं में अच्छी तरह सामान्यीकरण करते हैं और अर्जित जानकारियों को अच्छे से निकालते और जोड़ते हैं। खगोल-विज्ञान में, LLMs बुनियादी तथ्य आधारित सवालों में तो ठीक-ठाक काम करते हैं, लेकिन जब शब्दावली समझने, भौतिकी के नियम लागू करने या गणना वाले सवाल हल करने की बात आती है, तो इन्हें मुश्किल होती है। यानी, गणितीय तर्क और ज्ञानक्षेत्र डोमेन-विशिष्ट विशेषज्ञता में इनकी सीमाएं सामने आती हैं [2]। फिर भी, विज्ञान और अभियांत्रिकी में LLMs का इस्तेमाल बढ़ता जा रहा है। जैसे, German Space Operations Center (GSOC) ने अंतरिक्ष यान अभियान्त्रिकी में वास्तविक समय (real-time) पर समस्या निवारण, दस्तावेज़ीकरण और उचित निर्णय में समर्थन के लिए इनका उपयोग करना शुरू किया है [3]। इसी के साथ, AstroLLaMA [4] और astroBERT [5] जैसे प्रक्षेत्र- केंद्रित models ने तथ्यात्मक सटीकता, प्रासंगिक तर्क, और साहित्य प्रसंस्करण में अच्छा प्रदर्शन दिखाया है, जिससे यह साफ है कि लक्षित fine-tuning फायदेमंद है ।

इसके बावजूद, एक खामी है—अब तक ऐसा कोई विशेषीकृत LLM उपलब्ध नहीं है जो

खासतौर पर "अंतरिक्ष गतिविज्ञान" और "खगोलभौतिकी" के लिए तैयार किया गया हो, जहां वैचारिक समझ और गणितीय दृढ़ता दोनों जरूरी हैं। इसी कमी को पूरा करने के लिए हम प्रस्तावित करते हैं तारामंडल-GPT। यह एक domain-adapted फ्रेमवर्क है, जिसमें retrieval-based contextual grounding और structured thinking mechanisms को एकीकृत किया गया है। हमारा जोर factual accuracy और symbolic computation पर है, ताकि general-purpose assistance से आगे बढ़कर space science और engineering को आगे ले जाने के लिए भरोसेमंद tools तैयार किए जा सकें।

## 2. संबंधित कार्य

हाल के वर्षों में खगोल विज्ञान के लिए क्षेत्र-विशिष्ट कृत्रिम बुद्धिमत्ता (AI) में हुई प्रगति ने अनेक उल्लेखनीय मॉडलों को जन्म दिया है। इनमें AstroBERT [5] शामिल है, जो 110M parameter वाला BERT आधारित मॉडल है और जिसे खगोल-विज्ञान साहित्य पर प्रशिक्षित किया गया है ताकि उसका उपयोग समानार्थी खोज तथा सत्तात्मक पहचान में किया जा सके। इसी श्रेणी में AstroLLaMA परिवार उल्लेखनीय है, जहाँ हालिया अध्ययन में AstroLLaMA-2-7B आधारभूत मॉडल से 7–8% कमतर पाया गया, जबकि AstroLLaMA-3-8B ने ज्ञान संरक्षण किया पर astro-ph डेटा पर सुधार नहीं दिखाया। छोटे मॉडलों में catastrophic forgetting देखा गया, किन्तु AstroLLaMA-2-70B ने मूल LLaMA2-70B की तुलना में स्पष्ट लाभ दर्शाया, जिससे सिद्ध होता है कि खगोल—िशिष्ट सतत प्री-ट्रेनिंग का प्रभाव मुख्यतः 70B श्रेणी के मॉडलों पर ही उपयोगी है [4]। AstroMLab का AstroSage-LLaMA 3.1-8B [6] ने AstroMLab-1 पर 80.9% सटीकता प्राप्त की और अपेक्षाकृत कम पैरामीटरों के साथ GPT-4o के तुल्य प्रदर्शन किया। वहीं, बड़े AstroSage-LLaMA3.1-70B [7] मॉडल ने Claude-4Opus के साथ 86.2% के स्तर पर समान क्षमता प्रदर्शित की। इसके अतिरिक्त, StarWhisper LightCurve [8] श्रृंखला ने तार्किक वर्गीकरण में महत्वपूर्ण योगदान दिया है, जहाँ इसके Swin Transformer संस्करण ने 99% सटीकता प्राप्त की।

## 3. कार्यविधि/ कार्यप्रणाली

इस खंड में तारामंडल-GPT के विकास हेतु अपनाई गई कार्यप्रणाली का विवरण दिया गया है, जिसमें डाटासेट तैयारी, मॉडल चयन, Prompt इंजीनियरिंग, तथा fallback mechanism सहित retrieval-आधारित जनरेशन पाइपलाईन की अभिकल्पना सम्मिलित है।

### 3.1 डेटासेट विवरण एवं रूपांतरण प्रक्रिया

मॉडल ट्यूनिंग के लिए हमने open-source materials और पुस्तकें जैसे Fundamentals of Physics [9], Introduction to Space Physics [10], High-Energy Astrophysics [11] तथा संबंधित संदर्भों का उपयोग किया। Textual data को PyMuPdf python library की सहायता से निकाला गया और फिर इसे Markdown format में परिवर्तित किया गया, ताकि LLM model में retrieval के समय बेहतर समझ मिल सके। इसके बाद text को chunks में विभाजित किया गया, जहाँ प्रत्येक chunk size 1000 तथा chunk overlap 200 रखा गया, ताकि chunks के बीच संबंध बना रहे। इन chunks को Qwen3-Embedding-0.6B embedding model [12] की मदद से embed किया गया और embeddings को FAISS (Facebook Index Similarity Search) format में save किया गया। सहेजे गए FAISS vectorstore का उपयोग प्रश्नों के उत्तर देने के दौरान contextual information retrieval के लिए किया गया।.

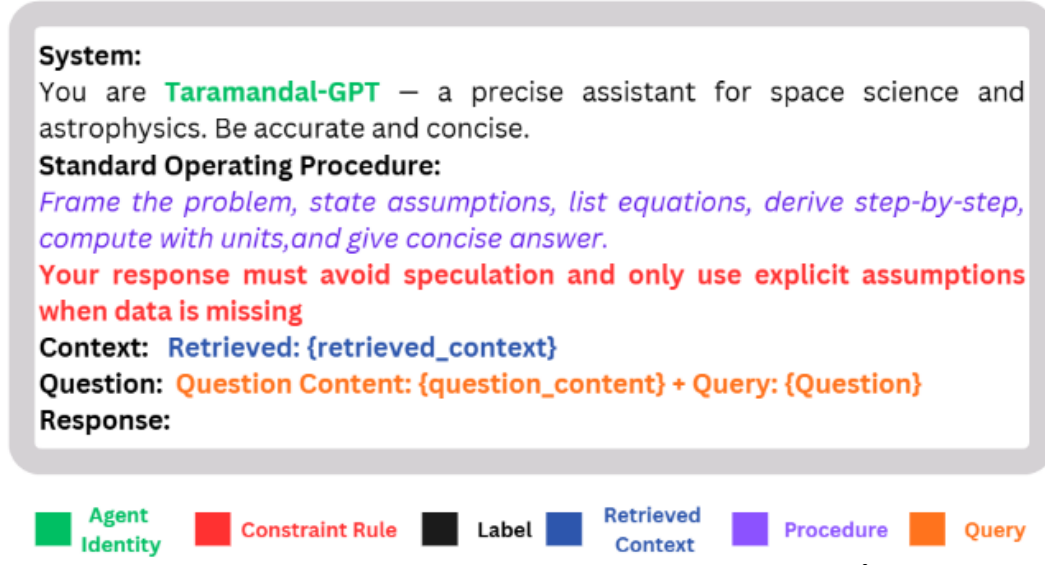

LLMs की खगोलगतिकी क्षेत्र में पूर्ण क्षमता का परीक्षण करने हेतु, हमने Astrodynamics Problems Benchmark (APBench) [13] का उपयोग किया है, जिसका उद्देश्य इस क्षेत्र में LLMs की क्षमताओं का आकलन करना है। इस benchmark में कुल 299 QA प्रश्न शामिल हैं, जिन्हें authoritative aerospace engineering foundational concepts से लेकर PhD-level तक की कठिनाई स्तरों को आच्छादित करते हैं।

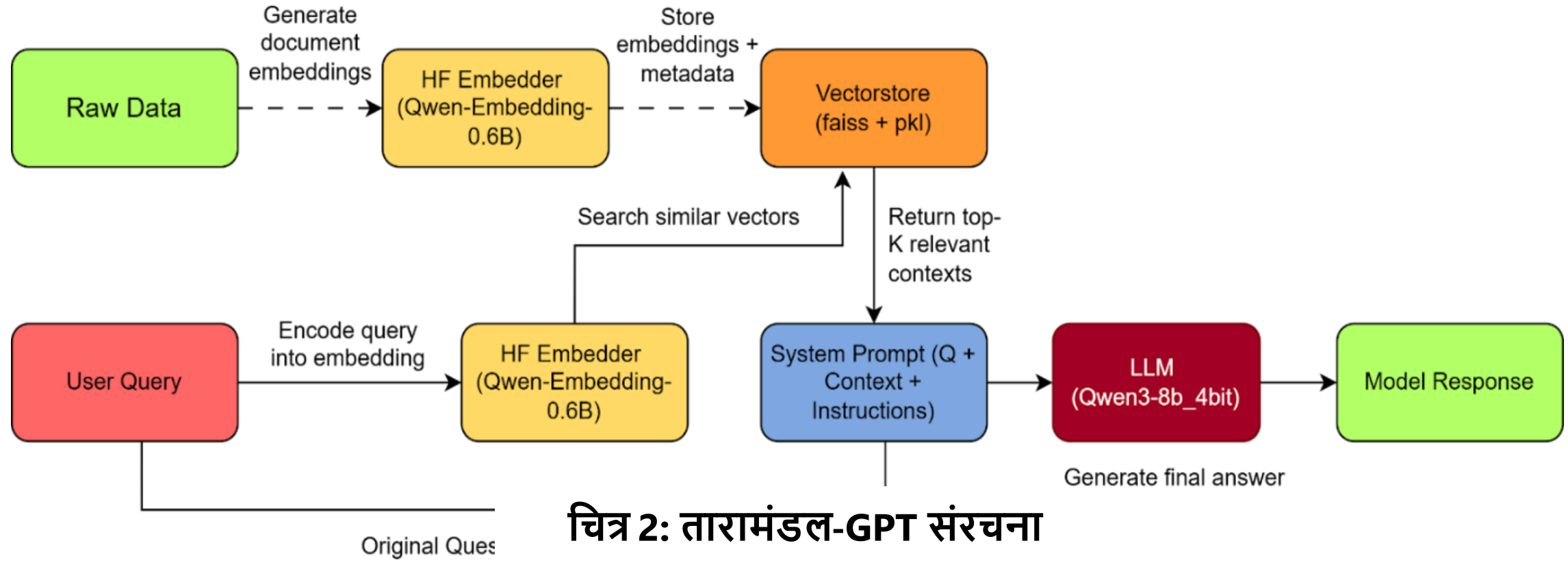

**चित्र 2: तारामंडल-GPT संरचना**

**तालिका 1: APBench प्रश्नों का स्तरवार वितरण**

| Benchmark Subset | Question Count |
|---|---|
| APBench – α | (gordon) 17 + (UBC) 8 |
| APBench – β | α (25) + (Braeunig) 144 |
| APBench – γ | β (169) + (lynnane) 130 |

यह dataset तीन क्रमिक स्तरों—alpha (α), beta (β) और gamma (γ)—में संरचित है, जिनका अनुपात तालिका 1 में प्रस्तुत किया गया है।

### 3.2 प्रस्तावित फ्रेमवर्क

प्रस्तावित फ्रेमवर्क , तारामंडल-GPT (जैसा कि चित्र 2 में दर्शाया गया है), में एक base pre-trained model, एक system prompt, तथा एक संबद्ध pipeline सम्मिलित है, जिनका विवरण निम्न उपखंडों में प्रस्तुत किया गया है।

**चित्र 1: System Prompt**

#### 3.1.1. बेस मॉडल (Base Model)

यह अध्ययन Qwen3-8b [14] का उपयोग करता है, जो एक खुला स्रोत LLM है और जटिल गणनाओं तथा ज्ञान-गहन कार्य में अपनी तर्क क्षमताओं के लिए प्रसिद्ध है। मॉडल को Q_4-बिट क्वांटाइजेशन के साथ परिनियोजित किया गया है ताकि हल्के निष्पादन और कुशल पहुंच सुनिश्चित की जा सके।

#### 3.1.2. सिस्टम प्रॉम्ट (System Prompt)

System prompt में chain-of-thought reasoning, surrogate feature utilization और regulatory mechanisms सम्मिलित हैं। ये विशेषताएँ मिलकर मॉडल को क्षेत्र-विशिष्ट उत्तरों के लिए fine-tune करती हैं, जिससे सटीकता में वृद्धि होती है। प्रणाली prompt का एक अवलोकन चित्र 1 में प्रस्तुत किया गया है।

#### 3.1.3. फ्रेमवर्क पाइपलाईन

प्रस्तावित फ्रेमवर्क, तारामंडल-GPT, एक Retrieval-Augmented Generation (RAG) आधारित वास्तुकला का उपयोग करता है, जिसे fallback mechanism के साथ सशक्त बनाया गया है। मुख्य कार्यप्रवाह में, प्रणाली बाहरी knowledge source से प्रासंगिक संदर्भ जानकारी (contextual data) पुनः प्राप्त करती है और उसे मॉडल की generative capabilities के साथ एकीकृत कर, सटीक और संदर्भ-सूचित उत्तर प्रदान करती है।

हालाँकि, जब पुनः प्राप्त संदर्भ विश्वसनीय परिणाम उत्पन्न करने के लिए अपर्याप्त या अधूरा होता है, तो fallback mechanism सक्रिय हो जाता है। इस स्थिति में, मॉडल अपने pre-trained knowledge base का उपयोग करता है और तर्क प्रक्रियाओं (reasoning processes) से गुजरते हुए पर्याप्त बाहरी डेटा की अनुपस्थिति में भी सटीक और संगत उत्तर निर्माण करता है।

संपूर्ण वास्तुशिल्प डिजाइन — जिसमें RAG pipeline और fallback pathway दोनों शामिल हैं — को चित्र 2 में दर्शाया गया है, जो प्रणाली के संचालन प्रवाह और निर्णय तर्क का स्पष्ट दृश्य प्रस्तुत करता है।

## 4. परिणाम एवं चर्चा

### 4.1. मॉडल मूल्यांकन मापदंड

APBench के लिए, मॉडल का प्रदर्शन अपेक्षित आउटपुट की प्रकृति के आधार पर, जो कि संख्यात्मक या संदेश-आधारित (message-based) हो सकता है, एक द्वि-फॉर्मेट स्कोरिंग स्कीमा द्वारा आंका जाता है।

#### 4.1.1. संख्यात्मक उत्तर का मूल्यांकन

संख्यात्मक उत्तरों का मूल सत्य के साथ मूल्यांकन एक dynamic error margin का उपयोग करके किया गया, जिसे समीकरण 1 में परिभाषित किया गया है।

$$\boldsymbol{Error_Margin} = \boldsymbol{min}(0.1 + 0.01 \cdot \boldsymbol{log}(|\, \boldsymbol{Answer} \,|), 0.1) \qquad \textbf{(1)}$$

यदि मॉडल का आउटपुट इस सीमा के भीतर आता है, तो उसे सही माना जाता है। उदाहरण के लिए, यदि वास्तविक उत्तर −5.2 है, तो स्वीकृत सीमा लगभग [−5.655, −4.479] होगी।

#### 4.1.2. संदेश-आधारित उत्तर का मूल्यांकन (Message Answer Scoring)

संदेश-आधारित उत्तरों का मूल्यांकन एक hybrid similarity approach द्वारा किया जाता है, जो semantic judgment और embedding-based comparison दोनों को सम्मिलित करता है। सबसे पहले, GPT-4o एक LLM-as-a-Judge के रूप में कार्य करता है और मॉडल-निर्मित उत्तर तथा संदर्भ उत्तर के बीच मेल के आधार पर 0 से 10 के बीच एक समानता स्कोर (similarity score) निर्धारित करता है। यह स्कोर बाद में सामान्यीकृत किया जाता है। समांतर रूप से, all-MiniLM-L6-v2 मॉडल से प्राप्त वाक्य एम्बेडिंग्स (sentence embeddings) का उपयोग कर cosine similarity की गणना की जाती है। अंतिम स्कोर इन दोनों घटकों के औसत के रूप में प्राप्त किया जाता है, जैसा कि समीकरण 2 में दिया गया है। सामान्यीकृत किया जाता है।

$$\boldsymbol{Score} = \boldsymbol{1/2} \times (\boldsymbol{LLM - Judge} + \boldsymbol{Embedding\ Similarity}) \qquad \textbf{(2)}$$

जो उत्तर 0.6 या उससे अधिक स्कोर प्राप्त करते हैं, उन्हें benchmark evaluation फ्रेमवर्क के अंतर्गत सटीक माना जाता है।

### 4.2 मूल्यांकन प्रोटोकॉल

पूरा मूल्यांकन प्रक्रिया में zero-shot prompting का उपयोग किया जाता है, जहाँ मॉडल्स को बिना किसी पूर्व उदाहरण या प्रदर्शनी के सीधे प्रश्न और संदर्भ प्रस्तुत किए जाते हैं।

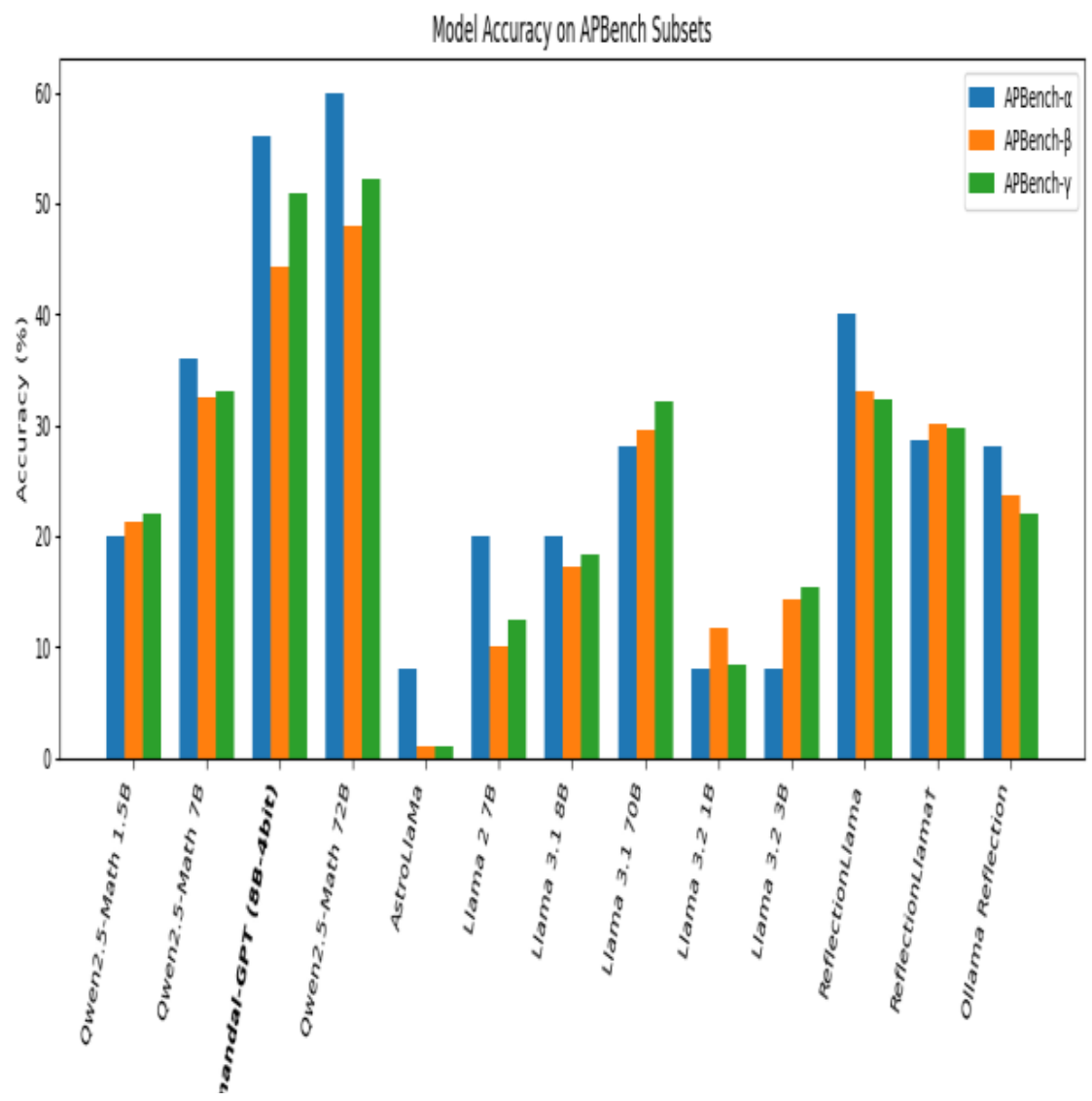

**चित्र 3: APBench पर तारामंडल-GPT का प्रदर्शन सहकर्मी खुला-स्रोत मॉडलों की तुलना में**

*स्रोत:* [https://www.nature.com/articles/s41598-025-91150-5](https://www.nature.com/articles/s41598-025-91150-5)

**4.3 तारामंडल-GPT का प्रदर्शन** प्रस्तावित तारामंडल-GPT का प्रदर्शन तालिका 3 में सारांशित किया गया है। इसके छोटे आकार और 4-bit Quantized होने के बावजूद, यह प्रदर्शन प्रतिस्पर्धात्मक है, विशेष रूप से बड़े मॉडलों जैसे Qwen2.5-Math 72B की तुलना में।

**तालिका 3: APBench के विभिन्न स्तरों पर तारामंडल-GPT का प्रदर्शन**

| APBench | α | β | γ |
|---|---|---|---|
| Accuracy (%) | 56 | 44.37 | 50.84 |

तारामंडल-GPT के प्रदर्शन की तुलना अन्य open-source मॉडलों से APBench-α, APBench-β, APBench-γ पर करते हैं। चित्र 3 में उल्लिखित विस्तृत मूल्यांकन परिणामों से प्राप्त मॉडल-वार प्रदर्शन प्रवृत्तियों का दृश्य प्रस्तुत किया गया है। तीनों APBench datasets पर व्यक्तिगत मॉडल के प्रदर्शन में कुछ विविधताएँ होने के बावजूद, open-source models के प्रदर्शन के बीच स्पष्ट अंतर दिखाई देता है।

**5. निष्कर्ष एवं भावी कार्यक्षेत्र**

इस कार्य में, हमने तारामंडल-GPT प्रस्तुत किया, जो खगोल विज्ञान और खगोलीय गतिशास्त्र में समस्या-समाधान को उन्नत करने हेतु एक क्षेत्र-विशिष्ट LLM फ्रेमवर्क है। Retrieval-Augmented Generation (RAG) पाइपलाइन को fallback mechanism के साथ संयोजित करके यह संदर्भ जागरूकता में सुधार करता है और सामान्य प्रयोजन तथा खगोल विज्ञान-केंद्रित मॉडलों की तुलना में अधिक विश्वसनीय उत्तर उत्पन्न करता है। APBench पर मूल्यांकन, जो खगोलीय गतिशास्त्र समस्याओं के लिए एक मानक है, ने इसके मौलिक सिद्धांतों से लेकर उन्नत अनुसंधान परिदृश्यों तक के कार्यों को संभालने की क्षमता दर्शाई है। ये परिणाम वैज्ञानिक क्षेत्रों में विशेषज्ञ LLMs की संभावनाओं को उजागर करते हैं, जहाँ सटीकता, गणितीय कठोरता, और व्याख्यात्मकता अत्यंत आवश्यक हैं, हालांकि hallucinations और धीमी विचार प्रक्रिया जैसी चुनौतियाँ बरकरार हुई हैं। नौ गुना कम पैरामीटर होने और 4 बिट क्वांटाइजेशन में क्वांटाइज्डहोने के बाद भी, मॉडल का परफॉर्मेंस बड़े 72B LLM के मुकाबले का है और इसमें सुदूर अंतरिक्ष के लिए कम ऊर्जा खपत तथा कम हार्डवेयर की जरुरत होती है।

आगे देखते हुए, भविष्य का कार्य प्रशिक्षण कॉर्पस का विस्तार करना है, जिसमें समकक्ष-परीक्षित साहित्य, उपग्रह टेलीमेट्री, और मिशन डिजाइन डेटा शामिल होंगे ताकि तथ्यात्मक आधार सुधारा जा सके; गणितीय विश्वसनीयता के लिए neuro-symbolic और physics-informed विधियों का समाकलन; और वैज्ञानिकों एवं अभियंताओं के साथ अंत:क्रियात्मक सहयोग को समर्थन देने के लिए एजेंट-आधारित विस्तार विकसित करना शामिल है। ग्रह विज्ञान, उपग्रह संचार, और बहिर्ग्रह मॉडलिंग जैसे अन्य क्षेत्रों में व्यापक मानकीकरण इस फ्रेमवर्क की मजबूती और प्रयोज्यता का और परीक्षण करेगा।

**6. संदर्भ**